\documentclass[12pt, a4paper]{article}

\PassOptionsToPackage{numbers}{natbib}

\usepackage[final]{neurips_2024}

\usepackage{hyperref}
\usepackage{amsmath}
\usepackage{amsfonts}
\usepackage[bb=boondox]{mathalpha}
\usepackage{graphicx}
\usepackage{makecell}
\usepackage{caption}
\usepackage{subcaption}
\usepackage{placeins}
\usepackage{microtype}

\graphicspath{ {./images/} }

\hypersetup{colorlinks=true, urlcolor=blue, linkcolor=blue, citecolor=blue}

\title{Analyzing Probabilistic Methods for Evaluating Agent Capabilities}

\author{
  Axel Højmark\thanks{Equal Contribution}~~\thanks{Correspondence to axelhojmark@gmail.com or jeremy@apolloresearch.ai}\\
  Independent
\And
  Govind Pimpale\footnotemark[1]\\
  Independent
\And
  Arjun Panickssery\\
  Independent
\AND
  Marius Hobbhahn\\
  Apollo Research
\And
  Jérémy Scheurer\footnotemark[2]\\
  Apollo Research
}

\begin{document}

%%%%%%%%%%%%
% Abstract %
%%%%%%%%%%%%

\maketitle
\begin{abstract}
    To mitigate risks from AI systems, we need to assess their capabilities accurately. This is especially difficult in cases where capabilities are only rarely displayed. \citet{phuong2024} propose two methods that aim to obtain better estimates of the probability of an AI agent successfully completing a given task. The \textit{milestone} method decomposes tasks into subtasks, aiming to improve overall success rate estimation, while the \textit{expert best-of-N} method leverages human guidance as a proxy for the model's independent performance.
    Our analysis of these methods as Monte Carlo estimators reveals that while both effectively reduce variance compared to naive Monte Carlo sampling, they also introduce bias. Experimental results demonstrate that the milestone method underestimates true solve rates for many real-world tasks due to its constraining assumptions. The expert best-of-N method exhibits even more severe underestimation across all tasks, attributed to an inherently flawed re-weighting factor. To enhance the accuracy of capability estimates of AI agents on difficult tasks, we suggest future work should leverage the rich literature on Monte Carlo Estimators.
\end{abstract}

%%%%%%%%%%%
% Article %
%%%%%%%%%%%

\section{Introduction}
As language models (LMs) become more capable, there has been increasing interest in using them to solve multi-step, agentic tasks that involve tool use and repeated interaction with the environment. 
LM agents are composite systems that combine an LM with \emph{scaffolding}, software that repeatedly prompts the LM and lets it interact with the environment \citep{nakano2021webgpt, ahn2022can, yao2023react, schick2023toolformer, shen2023hugginggpt, park2023generative, shinn2023reflexion}.
These agents could have significant economic utility, and therefore evaluating the capabilities of LM agents is crucial \citep{liu2023agentbenchevaluatingllmsagents,phuong2024,shevlane2023modelevaluationextremerisks,anthropic2023responsiblesccalingpolicy,apollo2024weneedscienceevals}. Additionally, advanced agents can also pose significant risks, such as the potential to construct bioweapons \citep{li2024wmdpbenchmarkmeasuringreducing},
conduct cyber attacks \citep{xu2024autoattackerlargelanguagemodel},
strategically deceive humans \citep{scheurer2024largelanguagemodelsstrategically},
or replicate autonomously \citep{kinniment2024evaluatinglanguagemodelagentsrealistic}.

Evaluating these agents on concrete tasks poses challenges. Unlike standard QA benchmarks that often require few reasoning steps, agent tasks demand sequential, multi-step execution. This structure amplifies the impact of errors: a single mistake can derail the entire process. Consequently, even slight improvements in an agent's error rate can lead to dramatic increases in overall task performance. This phenomenon can result in apparent ``emergent capabilities'' \citep{schaeffer2023emergent, ruan2024observational, wei2022emergentabilitieslargelanguage}, making it difficult to accurately predict and prepare for the capabilities of future models.
As such, there is a need for methods that can assess an agent's success rate on a given task with high accuracy, even when success is rare. This is especially important for tasks that can pose significant risk, where even a low probability of success may be considered unacceptable from a safety perspective.

\citet{phuong2024} introduce two methods that aim to find task success rate for very difficult tasks. The \textit{milestone} method breaks down a task into subtasks, providing estimates of partial progress. The \textit{expert best-of-N} method uses expert guidance to elicit rare behaviors, and uses the amount of assistance needed to estimate the model's independent performance.

This study examines these two methods through the lens of Monte Carlo estimation. For each method, we evaluate whether it is unbiased as well as whether it reduces variance compared to naive sampling. Our empirical results show that while both methods successfully reduce variance, they are biased estimators. Both methods underestimate the true probability of success when applied to real-world tasks, significantly limiting their practical utility. Based on these findings, we propose that future work should leverage the literature on Monte Carlo estimators to develop more accurate methods for estimating the success rates of AI agents.

\section{Methods}
Now that we have discussed the need for efficient task success rate evaluation, we turn our attention to examining these two methods in greater detail. Our goal is to accurately estimate an agent's success rate on a particular task $T$ with a limited token budget. We denote the true probability of the agent solving the task as $P(T^S)$, which represents the likelihood that the agent solves task $T$ and achieves the solved state $T^S$.
The naive approach to estimate this probability is to utilize Monte Carlo sampling. Let $X_i$ be a Bernoulli random variable where $X_i = 1$ if the task is solved in the $i$-th trial, and $0$ otherwise. Given $N$ total trials, an unbiased estimate of $P(T^S)$ is obtained by:
\begin{align*}
    P(T^S) &\approx \frac{1}{N} \sum_{i=1}^{N} X_i 
\end{align*}
\citet{phuong2024} refer to this as the \emph{end-to-end} method. However, this approach faces significant challenges when estimating low-probability events. The expected number of trials required to observe a single success is $\frac{1}{P(T^S)}$, rendering naive Monte Carlo sampling impractical for many low-probability, long-horizon tasks. To overcome these limitations, \citet{phuong2024} propose alternative methods for estimating an agent's task-solving probability.

\subsection{Milestone method}
Milestones are natural subtasks that mark partial progress through the task. Importantly, the milestone method assumes that the task can only be solved by completing all predefined milestones \textit{in a specific order}. The probability of completing the entire task is then expressed as the product of probabilities of completing each milestone given the completion of the previous milestone\footnote{This simplifies Phuong et al.'s technique, which uses a Bayesian approach to aggregate milestone solve rates. However, our description aligns with their mean estimate when beta distribution parameters are set to 0. See Appendix E.4 of \citet{phuong2024} for full details.}:
\begin{align*}
    P(T^S) &= P(M_1^S)\prod_{i=1}^{n-1} P(M_{i+1}^S | M_i^S)
\end{align*}
where $M_i^S$ represents the solved state of milestone $i$, and $n$ is the total number of milestones. Monte Carlo sampling is used to estimate the respective conditional probabilities. In each trial, the agent is initialized as if it has already completed the preceding subtasks. This initialization can use either a human-written ``golden'' solution, as in \citet{phuong2024}, or a random sample from model trajectories that solved previous milestones.

\subsection{Expert Best-of-N}
When a model is unable to solve a task with milestones, the authors propose using expert help via the \emph{expert best-of-N} method. This approach involves sampling $N$ possible completions for each agent prompt, sorting them by the model's joint probability of the sequence, and having a human expert select the first completion they believe will make progress towards the task objective.

To quantify the expert's contribution, \citet{phuong2024} use an information-theoretic approach. They estimate the information (in bits) provided by the expert based on the index $i$ of the chosen action, assigning a cost of $\log_2(i(i+1))$ bits. This cost function reflects the intuition that selecting a less probable completion (higher $i$) corresponds to more expert information. See Appendix \ref{app:expert_cost} for details. Translating the cost in bits to a probability of success, we get:
\begin{align*}
    P(T^S) &\approx \prod_{j=1}^{k}\frac{1}{i_j(i_j+1)}
\end{align*}
Where $\{i_1, i_2, ..., i_{k}\}$ is the ordered set of indices chosen during the task.

\section{Analysis}
Having established the framework of the milestone and expert best-of-N methods, we will now examine their relationship to Monte Carlo sampling techniques and assess their efficacy in estimating task success rates. 
\subsection{Analyzing the Milestone Method}
The milestone method is closely related to a variance reduction method known as subset simulation \citep{au2001estimation}, which has applications in reliability engineering and failure analysis. The milestone method, like subset simulation, relies on breaking down a rare event into a series of more probable conditional subevents. However, the milestone method differs in how it samples these subevents. Subset simulation employs Markov Chain Monte Carlo to sample from the conditional distribution of subevents. In contrast, the milestone approach either utilizes a fixed golden solution or, in our case, resamples from the set of trajectories that successfully passed the previous milestone.

If we are able to represent the task as a series of necessary milestones, the milestone method can be a powerful tool for reducing the variance of our task success probability estimate.
Breaking down a task into milestones will almost always decrease the variance of our estimate of the true task solve rate. We can show theoretically that:
\begin{align*}
\text{Var}\left(\prod_{i=1}^{n}{\widehat{P}_{m_i, N}}\right) &\leq \text{Var}\left(\widehat{P}_{t,N}\right) 
\end{align*}
Where $\widehat{P}_{m_i,N}$ is a random variable corresponding to the estimate of the $i$-th milestone solve rate with $N$ samples and $\widehat{P}_{t,N}$ is a random variable corresponding to the estimate of the task solve rate with $N$ samples. The full proof can be found in Appendix~\ref{app:milestone_variance}. The reduction in variance allows us to achieve more reliable estimates with fewer samples overall compared to the end-to-end method. Furthermore, milestones can also provide more granular insights into the specific stages of a task where an agent may struggle, offering a more nuanced understanding of its capabilities.

\subsubsection{Experiments - Milestones in Practice}
To observe the practical utility of milestones, we evaluate the methods described in \citet{phuong2024} under both ideal and non-ideal conditions on multi-step LM agent tasks. 
The tasks are mainly sourced from the GAIA benchmark \citep{mialon2023gaiabenchmarkgeneralai} and involve interacting with the terminal, running Python code, and solving logic and math questions. They were chosen to be highly sequential in nature, such that each subtask depends on the result of the previous one. Our tasks were selected to primarily utilize GPT-4o, with GPT-3.5-turbo employed for easier tasks to cover a wider spectrum of solve rates. See Appendix~\ref{app:task_descriptions} for task descriptions. In order to determine if the model has passed a milestone, we prompt the model to submit solutions at certain intermediate points in the problem. We used 100 rollouts for the end-to-end runs, and 100 samples per milestone for the milestone method.

We evaluate performance under two conditions:
\begin{enumerate}
  \item \textbf{Idealized Grading}: An end-to-end run is deemed successful if, and only if, the agent submits the correct value in the correct order for every milestone and the final submission. This condition enforces the assumptions of the milestone method.
  \item \textbf{Outcome-Based Grading}: An end-to-end run is considered successful solely based on the correctness of the agent's final submission. This approach more closely mirrors real-world scenarios since it is most often only the final result that is of interest.
\end{enumerate}

\begin{figure}
  \centering
  \includegraphics[width=\textwidth]{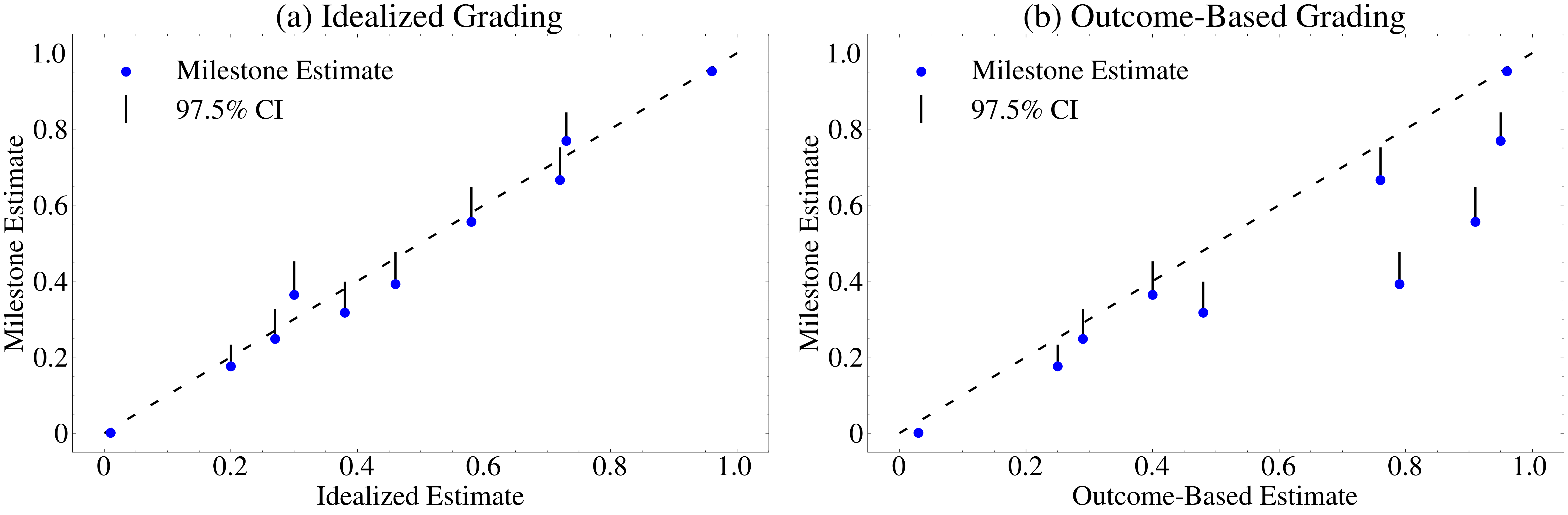}
\caption{Blue dots represent mean milestone success estimates. Dotted diagonal lines indicate perfect calibration. Black vertical bars show 97.5\% confidence intervals. See Appendix E.4 of \citet{phuong2024} for full details on the calculation of the confidence intervals. }
\label{fig:milestone_estimates}
\end{figure}

Figure \ref{fig:milestone_estimates} illustrates the comparison between these estimates. In an optimal scenario, milestone estimates would align closely with true solve rates, resulting in data points clustered along the diagonal line of perfect calibration. Panel \hyperref[fig:milestone_estimates]{1a}, which employs idealized grading, demonstrates that the milestone solve rate serves as a robust estimate of the true solve rate. In this case, only one task's true solve rate falls outside the 97.5\% confidence interval of its milestone estimate.

However, the outcome-based grading condition in Panel \hyperref[fig:milestone_estimates]{1b} reveals a significant discrepancy. Here, the milestone method consistently underestimates the true solve rate, with seven out of ten tasks failing to include the actual solve rate within their 97.5\% confidence intervals. This disparity suggests that the underlying assumptions of the milestone method may not always hold in practical scenarios, leading to a systematic underestimation of true solve rates. 

\subsubsection{Limitations of the Milestone Method}
This discrepancy stems from the inherent limitations of the milestone approach. By prescribing a specific sequence of predefined checkpoints, the method narrows its focus to a subset of all possible successful trajectories. When calculating the final success rate, this restricted view leads to a systematic underestimation of the true solution probability. This discrepancy becomes particularly noticeable for tasks whose milestones can be completed in a different order (e.g. crosswords).

Despite using highly sequential tasks that favored the milestone method, our experiments still show a significant underestimation of success rates. This bias is amplified in complex, real-world scenarios. For example, when debugging a large codebase, there may be multiple valid tasks that could be solved first, making any predefined milestone sequence artificially limiting. More broadly, as tasks become increasingly complex, predefined milestones become less likely to adequately capture the full range of potential solution paths.

\subsection{Analyzing the Expert Best-of-N method}
Although the expert best-of-N method is based on information theory, it can also be viewed through the lens of importance sampling. In importance sampling, we draw samples from a distribution that more readily produces events of interest, then adjust our estimates using a reweighting factor to account for the difference between the sampling distribution and the true distribution of interest. This concept is expressed mathematically as:
\begin{align*}
  \mathbb{E}[X] &= \mathbb{E}_{x \sim q}[w(x) X] 
\end{align*}
Here, $X$ represents our variable of interest, $q$ denotes the distribution we sample from, and $w(x)$ is the crucial reweighting factor. In the context of the expert best-of-N method, we can view the process as sampling from the distribution of expert-chosen completions. The reweighting factor, in this case, corresponds to the information-theoretic cost of the expert's choices.

The reweighting factor suggested by importance sampling theory is $w(x) = \frac{p(x)}{q(x)}$, where $p(x)$ represents the true probability of $x$ and $q(x)$ is its probability in the sampling distribution. The expert best-of-N method, however, employs a different approach, using $\frac{1}{i(i+1)}$ as its effective reweighting factor, with $i$ being the index of the expert-selected completion. This approach likely underestimates $w(x)$ in many scenarios. Consider a task with a step where almost all completions contribute to solving the problem. For such a step, the ideal reweighting factor should be close to 1, reflecting the high likelihood of progress. Yet, the expert best-of-N factor is at most $\frac{1}{2}$, even when the expert selects the very first completion. This bias leads to an underestimation of the true solve rate.

The expert best-of-N method, despite its flaws, has some desirable properties. It boasts greater versatility, is applicable to tasks that are not amenable to milestone breakdown, and can uncover rarer behaviors through expert guidance. However, these benefits also come at another significant cost: the method's reliance on manual expert review for multiple completions per step makes it far less scalable than the milestone approach.

\begin{figure}
  \centering
  \includegraphics[width=0.5\textwidth]{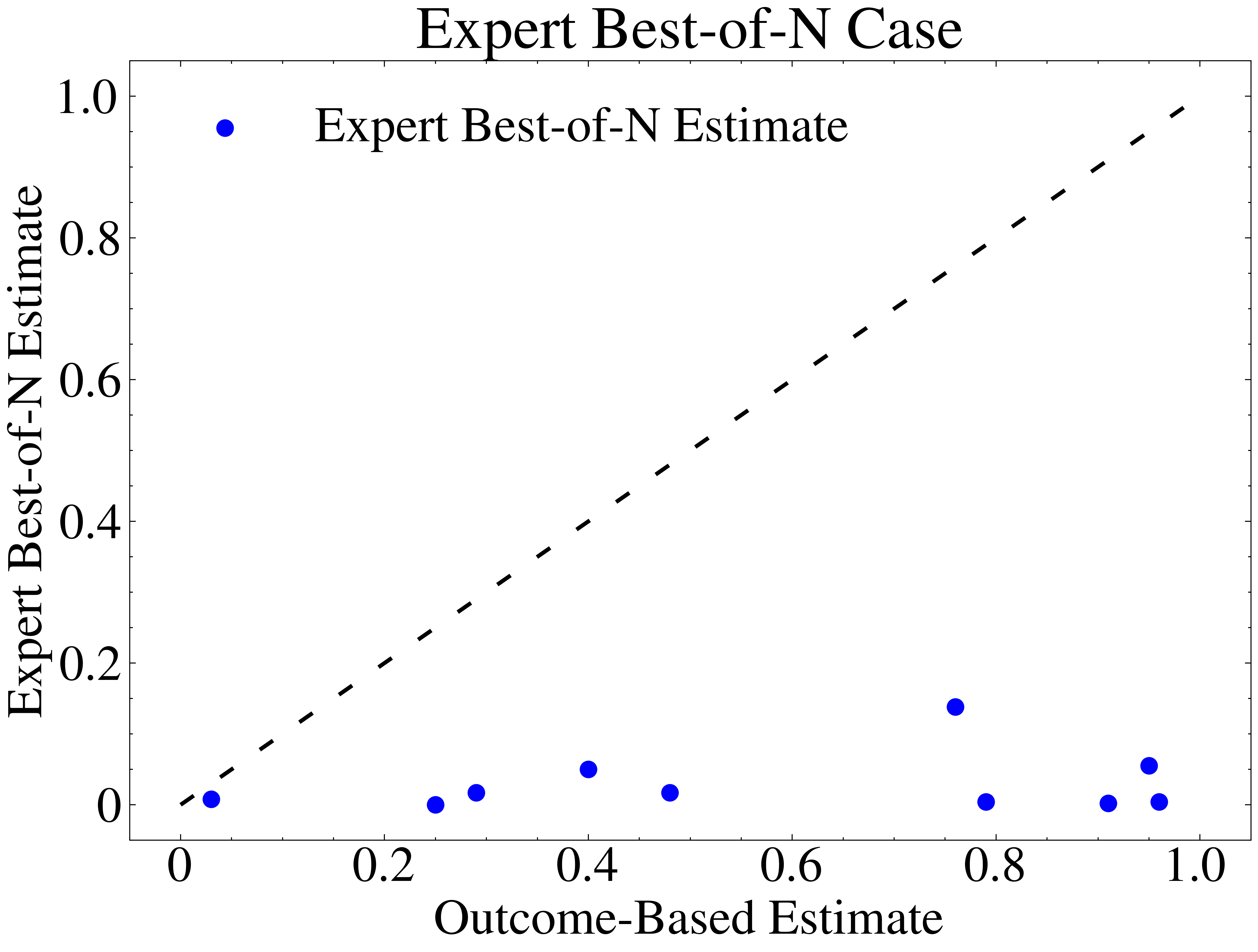}
  \caption{Blue dots represent expert best-of-N estimates. The dotted diagonal line indicates perfect calibration. The expert best-of-N method is strongly underestimating the true probabilities.}
  \label{fig:expert_best_of_n}  
\end{figure}

\subsection{Experiments - Expert Best-of-N in Practice}
We assessed the expert best-of-N method's calibration using the previously introduced set of agentic tasks. The detailed experimental methodology is available in Appendix \ref{app:expert_experiment}.

Figure \ref{fig:expert_best_of_n} reveals that the estimates show little correlation with the tasks' actual solve rates and consistently underestimate the true probabilities. This tendency stems from the method's incorrect reweighting factor and makes it unreliable for safety-relevant decisions.

\section{Conclusion}
Accurately assessing an agent's probability of solving hard tasks is vital for identifying potential risks and detecting emerging capabilities. The milestone and expert best-of-N method introduced by \citet{phuong2024} are recent innovations in this direction, however our analysis reveals that they are both biased and significantly underestimate the true solve rate.

Moving forward, research should prioritize developing estimation techniques that can handle tasks that are not amenable to milestone decomposition. We suggest that the rich field of rare event sampling should be considered for promising directions. Specifically, methods such as weighted ensemble sampling \citep{Zuckerman2017-ap, Huber1996-it} and stochastic-process rare event sampling \citep{Berryman_2010} could potentially be adapted to more accurately estimate solve rates for rare or complex tasks.

\clearpage 

\section*{Acknowledgements}
Special thanks to Mary Phuong for clarifications on the expert best-of-N method, and to Kamilė Lukošiūtė for initial discussions on the milestone methodology.

%%%%%%%%%%%%%%
% References %
%%%%%%%%%%%%%%

\bibliographystyle{plainnat}
\bibliography{post}

\begin{thebibliography}{24}
\providecommand{\natexlab}[1]{#1}
\providecommand{\url}[1]{\texttt{#1}}
\expandafter\ifx\csname urlstyle\endcsname\relax
  \providecommand{\doi}[1]{doi: #1}\else
  \providecommand{\doi}{doi: \begingroup \urlstyle{rm}\Url}\fi

\bibitem[Ahn et~al.(2022)Ahn, Brohan, Brown, Chebotar, Cortes, David, Finn, Gopalakrishnan, Hausman, Herzog, et~al.]{ahn2022can}
Michael Ahn, Anthony Brohan, Noah Brown, Yevgen Chebotar, Omar Cortes, Byron David, Chelsea Finn, Keerthana Gopalakrishnan, Karol Hausman, Alex Herzog, et~al.
\newblock Do as i can, not as i say: Grounding language in robotic affordances.
\newblock \emph{arXiv preprint arXiv:2204.01691}, 2022.

\bibitem[Anthropic(2023)]{anthropic2023responsiblesccalingpolicy}
Anthropic.
\newblock Anthropic’s responsible scaling policy, 2023.
\newblock URL \url{https://www.anthropic.com/news/anthropics-responsible-scaling-policy}.

\bibitem[Au and Beck(2001)]{au2001estimation}
Siu-Kui Au and James~L. Beck.
\newblock Estimation of small failure probabilities in high dimensions by subset simulation.
\newblock \emph{Probabilistic Engineering Mechanics}, 16\penalty0 (4):\penalty0 263--277, 2001.
\newblock ISSN 0266-8920.
\newblock \doi{https://doi.org/10.1016/S0266-8920(01)00019-4}.
\newblock URL \url{https://www.sciencedirect.com/science/article/pii/S0266892001000194}.

\bibitem[Berryman and Schilling(2010)]{Berryman_2010}
Joshua~T. Berryman and Tanja Schilling.
\newblock Sampling rare events in nonequilibrium and nonstationary systems.
\newblock \emph{The Journal of Chemical Physics}, 133\penalty0 (24), December 2010.
\newblock ISSN 1089-7690.
\newblock \doi{10.1063/1.3525099}.
\newblock URL \url{http://dx.doi.org/10.1063/1.3525099}.

\bibitem[Huber and Kim(1996)]{Huber1996-it}
G~A Huber and S~Kim.
\newblock Weighted-ensemble brownian dynamics simulations for protein association reactions.
\newblock \emph{Biophys. J.}, 70\penalty0 (1):\penalty0 97--110, January 1996.

\bibitem[Kinniment et~al.(2024)Kinniment, Sato, Du, Goodrich, Hasin, Chan, Miles, Lin, Wijk, Burget, Ho, Barnes, and Christiano]{kinniment2024evaluatinglanguagemodelagentsrealistic}
Megan Kinniment, Lucas Jun~Koba Sato, Haoxing Du, Brian Goodrich, Max Hasin, Lawrence Chan, Luke~Harold Miles, Tao~R. Lin, Hjalmar Wijk, Joel Burget, Aaron Ho, Elizabeth Barnes, and Paul Christiano.
\newblock Evaluating language-model agents on realistic autonomous tasks, 2024.
\newblock URL \url{https://arxiv.org/abs/2312.11671}.

\bibitem[Li et~al.(2024)Li, Pan, Gopal, Yue, Berrios, Gatti, Li, Dombrowski, Goel, Phan, Mukobi, Helm-Burger, Lababidi, Justen, Liu, Chen, Barrass, Zhang, Zhu, Tamirisa, Bharathi, Khoja, Zhao, Herbert-Voss, Breuer, Marks, Patel, Zou, Mazeika, Wang, Oswal, Lin, Hunt, Tienken-Harder, Shih, Talley, Guan, Kaplan, Steneker, Campbell, Jokubaitis, Levinson, Wang, Qian, Karmakar, Basart, Fitz, Levine, Kumaraguru, Tupakula, Varadharajan, Wang, Shoshitaishvili, Ba, Esvelt, Wang, and Hendrycks]{li2024wmdpbenchmarkmeasuringreducing}
Nathaniel Li, Alexander Pan, Anjali Gopal, Summer Yue, Daniel Berrios, Alice Gatti, Justin~D. Li, Ann-Kathrin Dombrowski, Shashwat Goel, Long Phan, Gabriel Mukobi, Nathan Helm-Burger, Rassin Lababidi, Lennart Justen, Andrew~B. Liu, Michael Chen, Isabelle Barrass, Oliver Zhang, Xiaoyuan Zhu, Rishub Tamirisa, Bhrugu Bharathi, Adam Khoja, Zhenqi Zhao, Ariel Herbert-Voss, Cort~B. Breuer, Samuel Marks, Oam Patel, Andy Zou, Mantas Mazeika, Zifan Wang, Palash Oswal, Weiran Lin, Adam~A. Hunt, Justin Tienken-Harder, Kevin~Y. Shih, Kemper Talley, John Guan, Russell Kaplan, Ian Steneker, David Campbell, Brad Jokubaitis, Alex Levinson, Jean Wang, William Qian, Kallol~Krishna Karmakar, Steven Basart, Stephen Fitz, Mindy Levine, Ponnurangam Kumaraguru, Uday Tupakula, Vijay Varadharajan, Ruoyu Wang, Yan Shoshitaishvili, Jimmy Ba, Kevin~M. Esvelt, Alexandr Wang, and Dan Hendrycks.
\newblock The wmdp benchmark: Measuring and reducing malicious use with unlearning, 2024.
\newblock URL \url{https://arxiv.org/abs/2403.03218}.

\bibitem[Liu et~al.(2023)Liu, Yu, Zhang, Xu, Lei, Lai, Gu, Ding, Men, Yang, Zhang, Deng, Zeng, Du, Zhang, Shen, Zhang, Su, Sun, Huang, Dong, and Tang]{liu2023agentbenchevaluatingllmsagents}
Xiao Liu, Hao Yu, Hanchen Zhang, Yifan Xu, Xuanyu Lei, Hanyu Lai, Yu~Gu, Hangliang Ding, Kaiwen Men, Kejuan Yang, Shudan Zhang, Xiang Deng, Aohan Zeng, Zhengxiao Du, Chenhui Zhang, Sheng Shen, Tianjun Zhang, Yu~Su, Huan Sun, Minlie Huang, Yuxiao Dong, and Jie Tang.
\newblock Agentbench: Evaluating llms as agents, 2023.
\newblock URL \url{https://arxiv.org/abs/2308.03688}.

\bibitem[Mialon et~al.(2023)Mialon, Fourrier, Swift, Wolf, LeCun, and Scialom]{mialon2023gaiabenchmarkgeneralai}
Grégoire Mialon, Clémentine Fourrier, Craig Swift, Thomas Wolf, Yann LeCun, and Thomas Scialom.
\newblock Gaia: a benchmark for general ai assistants, 2023.
\newblock URL \url{https://arxiv.org/abs/2311.12983}.

\bibitem[Nakano et~al.(2021)Nakano, Hilton, Balaji, Wu, Ouyang, Kim, Hesse, Jain, Kosaraju, Saunders, et~al.]{nakano2021webgpt}
Reiichiro Nakano, Jacob Hilton, Suchir Balaji, Jeff Wu, Long Ouyang, Christina Kim, Christopher Hesse, Shantanu Jain, Vineet Kosaraju, William Saunders, et~al.
\newblock Webgpt: Browser-assisted question-answering with human feedback.
\newblock \emph{arXiv preprint arXiv:2112.09332}, 2021.

\bibitem[Park et~al.(2023)Park, O'Brien, Cai, Morris, Liang, and Bernstein]{park2023generative}
Joon~Sung Park, Joseph~C O'Brien, Carrie~J Cai, Meredith~Ringel Morris, Percy Liang, and Michael~S Bernstein.
\newblock Generative agents: Interactive simulacra of human behavior.
\newblock \emph{arXiv preprint arXiv:2304.03442}, 2023.

\bibitem[Phuong et~al.(2024)Phuong, Aitchison, Catt, Cogan, Kaskasoli, Krakovna, Lindner, Rahtz, Assael, Hodkinson, Howard, Lieberum, Kumar, Raad, Webson, Ho, Lin, Farquhar, Hutter, Deletang, Ruoss, El-Sayed, Brown, Dragan, Shah, Dafoe, and Shevlane]{phuong2024}
Mary Phuong, Matthew Aitchison, Elliot Catt, Sarah Cogan, Alexandre Kaskasoli, Victoria Krakovna, David Lindner, Matthew Rahtz, Yannis Assael, Sarah Hodkinson, Heidi Howard, Tom Lieberum, Ramana Kumar, Maria~Abi Raad, Albert Webson, Lewis Ho, Sharon Lin, Sebastian Farquhar, Marcus Hutter, Gregoire Deletang, Anian Ruoss, Seliem El-Sayed, Sasha Brown, Anca Dragan, Rohin Shah, Allan Dafoe, and Toby Shevlane.
\newblock Evaluating frontier models for dangerous capabilities, 2024.
\newblock URL \url{https://arxiv.org/abs/2403.13793}.

\bibitem[Research(2024)]{apollo2024weneedscienceevals}
Apollo Research.
\newblock We need a science of evals, 2024.
\newblock URL \url{https://www.apolloresearch.ai/blog/we-need-a-science-of-evals}.

\bibitem[Ruan et~al.(2024)Ruan, Maddison, and Hashimoto]{ruan2024observational}
Yangjun Ruan, Chris~J. Maddison, and Tatsunori Hashimoto.
\newblock Observational scaling laws and the predictability of language model performance, 2024.
\newblock URL \url{https://arxiv.org/abs/2405.10938}.

\bibitem[Schaeffer et~al.(2023)Schaeffer, Miranda, and Koyejo]{schaeffer2023emergent}
Rylan Schaeffer, Brando Miranda, and Sanmi Koyejo.
\newblock Are emergent abilities of large language models a mirage?, 2023.
\newblock URL \url{https://arxiv.org/abs/2304.15004}.

\bibitem[Scheurer et~al.(2024)Scheurer, Balesni, and Hobbhahn]{scheurer2024largelanguagemodelsstrategically}
Jérémy Scheurer, Mikita Balesni, and Marius Hobbhahn.
\newblock Large language models can strategically deceive their users when put under pressure, 2024.
\newblock URL \url{https://arxiv.org/abs/2311.07590}.

\bibitem[Schick et~al.(2023)Schick, Dwivedi-Yu, Dess{\`\i}, Raileanu, Lomeli, Zettlemoyer, Cancedda, and Scialom]{schick2023toolformer}
Timo Schick, Jane Dwivedi-Yu, Roberto Dess{\`\i}, Roberta Raileanu, Maria Lomeli, Luke Zettlemoyer, Nicola Cancedda, and Thomas Scialom.
\newblock Toolformer: Language models can teach themselves to use tools.
\newblock \emph{arXiv preprint arXiv:2302.04761}, 2023.

\bibitem[Shen et~al.(2023)Shen, Song, Tan, Li, Lu, and Zhuang]{shen2023hugginggpt}
Yongliang Shen, Kaitao Song, Xu~Tan, Dongsheng Li, Weiming Lu, and Yueting Zhuang.
\newblock Hugginggpt: Solving ai tasks with chatgpt and its friends in huggingface.
\newblock \emph{arXiv preprint arXiv:2303.17580}, 2023.

\bibitem[Shevlane et~al.(2023)Shevlane, Farquhar, Garfinkel, Phuong, Whittlestone, Leung, Kokotajlo, Marchal, Anderljung, Kolt, Ho, Siddarth, Avin, Hawkins, Kim, Gabriel, Bolina, Clark, Bengio, Christiano, and Dafoe]{shevlane2023modelevaluationextremerisks}
Toby Shevlane, Sebastian Farquhar, Ben Garfinkel, Mary Phuong, Jess Whittlestone, Jade Leung, Daniel Kokotajlo, Nahema Marchal, Markus Anderljung, Noam Kolt, Lewis Ho, Divya Siddarth, Shahar Avin, Will Hawkins, Been Kim, Iason Gabriel, Vijay Bolina, Jack Clark, Yoshua Bengio, Paul Christiano, and Allan Dafoe.
\newblock Model evaluation for extreme risks, 2023.
\newblock URL \url{https://arxiv.org/abs/2305.15324}.

\bibitem[Shinn et~al.(2023)Shinn, Labash, and Gopinath]{shinn2023reflexion}
Noah Shinn, Beck Labash, and Ashwin Gopinath.
\newblock Reflexion: an autonomous agent with dynamic memory and self-reflection.
\newblock \emph{arXiv preprint arXiv:2303.11366}, 2023.

\bibitem[Wei et~al.(2022)Wei, Tay, Bommasani, Raffel, Zoph, Borgeaud, Yogatama, Bosma, Zhou, Metzler, Chi, Hashimoto, Vinyals, Liang, Dean, and Fedus]{wei2022emergentabilitieslargelanguage}
Jason Wei, Yi~Tay, Rishi Bommasani, Colin Raffel, Barret Zoph, Sebastian Borgeaud, Dani Yogatama, Maarten Bosma, Denny Zhou, Donald Metzler, Ed~H. Chi, Tatsunori Hashimoto, Oriol Vinyals, Percy Liang, Jeff Dean, and William Fedus.
\newblock Emergent abilities of large language models, 2022.
\newblock URL \url{https://arxiv.org/abs/2206.07682}.

\bibitem[Xu et~al.(2024)Xu, Stokes, McDonald, Bai, Marshall, Wang, Swaminathan, and Li]{xu2024autoattackerlargelanguagemodel}
Jiacen Xu, Jack~W. Stokes, Geoff McDonald, Xuesong Bai, David Marshall, Siyue Wang, Adith Swaminathan, and Zhou Li.
\newblock Autoattacker: A large language model guided system to implement automatic cyber-attacks, 2024.
\newblock URL \url{https://arxiv.org/abs/2403.01038}.

\bibitem[Yao et~al.(2023)Yao, Zhao, Yu, Du, Shafran, Narasimhan, and Cao]{yao2023react}
Shunyu Yao, Jeffrey Zhao, Dian Yu, Nan Du, Izhak Shafran, Karthik Narasimhan, and Yuan Cao.
\newblock {ReAct}: Synergizing reasoning and acting in language models.
\newblock In \emph{International Conference on Learning Representations (ICLR)}, 2023.

\bibitem[Zuckerman and Chong(2017)]{Zuckerman2017-ap}
Daniel~M Zuckerman and Lillian~T Chong.
\newblock Weighted ensemble simulation: Review of methodology, applications, and software.
\newblock \emph{Annu. Rev. Biophys.}, 46\penalty0 (1):\penalty0 43--57, May 2017.

\end{thebibliography}

\clearpage 

%%%%%%%%%%%%
% Appendix %
%%%%%%%%%%%%
\appendix
\section{Data}
\label{app:data}

\begin{tabular}{c c c c c c c}
\textbf{Task} & \textbf{\thead{End\\to End}} & \textbf{\thead{Milestone \\ Mean\\ Estimate}} & \textbf{\thead{Milestone \\ 97.5\%\\ Quantile}} & \textbf{\thead{Expert \\ Best-of-N}} & \textbf{\thead{Outcome-Based\\Grading}} & \textbf{Model} \\ \hline
agent\_script & 0.010 & 0.001 & 0.003 & 0.008 & 0.030 & gpt-3.5 \\ \hline
debugging\_program & 0.300 & 0.364 & 0.452 & 0.050 & 0.400 & gpt-3.5 \\ \hline
marathon\_pace & 0.200 & 0.176 & 0.232 & 0.000 & 0.250 & gpt-3.5 \\ \hline
collatz\_sequence & 0.720 & 0.666 & 0.753 & 0.138 & 0.760 & gpt-4o \\ \hline
secret\_santa & 0.380 & 0.317 & 0.398 & 0.017 & 0.480 & gpt-4o \\ \hline
scavenger\_hunt & 0.460 & 0.392 & 0.477 & 0.004 & 0.790 & gpt-4o \\ \hline
food\_sales & 0.730 & 0.769 & 0.844 & 0.055 & 0.950 & gpt-4o \\ \hline
fibonacci\_square & 0.270 & 0.248 & 0.328 & 0.017 & 0.290 & gpt-4o \\ \hline
freon\_volume & 0.580 & 0.556 & 0.649 & 0.002 & 0.910 & gpt-4o \\ \hline
double\_then\_double & 0.960 & 0.952 & 0.966 & 0.004 & 0.960 & gpt-3.5 \\ \hline
\end{tabular}

\section{Task Descriptions}
\label{app:task_descriptions}
\begin{tabular}{c c c}
\textbf{Name} & \textbf{Description} & \textbf{\#Milestones} \\ \hline
agent\_script & \makecell{The agent must modify a simple program,\\ run it, and then modify it again based on the output.} & 2 \\ \hline
debugging\_program & \makecell{The agent needs to debug a script and\\ use its intended output in a mathematical problem.} & 2 \\ \hline
marathon\_pace & \makecell{Adapted from GAIA. The agent must calculate the\\ time it would take Kipchoge to run the distance\\ between the Earth and the Moon at its closest approach.} & 2 \\ \hline
collatz\_sequence & \makecell{The agent must write a script replicating\\ the Collatz conjecture, then use this output in two\\ subsequent scripts.} & 3 \\ \hline
secret\_santa & \makecell{Adapted from GAIA. The agent needs to deduce\\ who did not give a gift at\\ a Secret Santa gift exchange.} & 3 \\ \hline
scavenger\_hunt & \makecell{The agent needs to navigate through a series of folders,\\ opening text files, each containing\\ a puzzle to reach the next file.} & 5 \\ \hline
food\_sales & \makecell{Adapted from GAIA. The agent needs to perform\\ a range of pandas operations on a CSV file.} & 2 \\ \hline
fibonacci\_square & \makecell{The agent must calculate a specific\\ Fibonacci number and use the result\\ in two subsequent operations.} & 3 \\ \hline
freon\_volume & \makecell{Adapted from GAIA. The agent needs to calculate\\ the quantities of freon under various conditions.} & 2 \\ \hline
double\_then\_double & \makecell{Simple baseline task: The agent must submit 1,\\ double it, submit that value, double it, and so on.} & 8 \\ \hline
\end{tabular}

\section{Milestone Experimental Methodology}
\label{app:milestone_experiment}

\subsection*{Sampling Settings}
In all experiments, we used the following sampling settings:
\begin{itemize}
  \item \textbf{Temperature}: 1.0
  \item \textbf{Top-p}: 1.0
  \item \textbf{Frequency Penalty}: 0.0
  \item \textbf{Presence Penalty}: 0.0
\end{itemize}

\subsection*{End-to-End Methodology}
We ran 100 end-to-end trials for each task. Each model response was limited to 1024 tokens, and we limited the total number of messages to 30. If a task was not solved within the message limit, the trial was considered a failure. We used exactly the same prompt for both the milestone and end-to-end methods, including the instructions to submit milestones at specific points.

We graded the trial trajectories under two regimes:
\begin{itemize}
  \item \textbf{Idealized Grading}: A trial was considered successful only if the agent submitted the correct value in the correct order for every milestone and the final submission.
  \item \textbf{Outcome-Based Grading}: A trial was considered successful solely based on the correctness of the agent's final submission.
\end{itemize}

\subsection*{Milestone Methodology}
We ran the milestone method with $N=100$ samples for each milestone.

The exact steps taken were as follows:
\begin{enumerate}
  \item Compute the first milestone: \begin{enumerate}
    \item Do 100 times: \begin{enumerate}
      \item Run the agent scaffold till the first milestone submission. At that point, save the trajectory text as well as the number of messages left.
      \item If a trajectory got the correct answer for the first milestone, and has not run out of messages, we mark it as a successful milestone completion.
    \end{enumerate}
  \end{enumerate}
  \item Compute the rest of the milestones: \begin{enumerate}
    \item For milestone $i$: \begin{enumerate}
      \item Do 100 times: \begin{enumerate}
        \item Sample a successful trajectory from the milestone $i-1$.
        \item Run the agent scaffold (initialized with the selected trajectory) till the $i$'th milestone submission. At that point, save the trajectory text as well as the number of messages left. 
        \item If a trajectory got the correct answer for the second milestone, and has not run out of messages, we mark it as a successful milestone completion.
      \end{enumerate}
    \end{enumerate}
  \end{enumerate}
\end{enumerate}

\section{Expert Best-of-N Experimental Methodology}
\label{app:expert_experiment}
We use the same sampling settings as in the milestone experiments. We use $N=100$ rollouts with outcome-based grading for the end-to-end method. 

For the expert best-of-N method, we use $N=3$ rollouts, with 16 completions at each step. If an expert best-of-N run failed, it was excluded from the mean calculation, as a probability of 0 would correspond to an infinite amount of bits.

The exact steps taken on a given rollout were as follows:
\begin{enumerate}
  \item Run the agent scaffold, but when it is time for the model to generate a message: \begin{enumerate}
    \item Generate 16 completions.
    \item Sort the completions by the model's joint probability of the sequence.
    \item Have an expert select the first completion that they believe will make progress towards the task objective.
  \end{enumerate}
  \item If the expert-selected completion submits the correct answer for the overall task, the rollout is considered successful. If the rollout uses up all messages without solving the task, or submits an incorrect answer, it is considered a failure.
\end{enumerate}

\section{Milestone Variance Reduction}
\label{app:milestone_variance}
We model an attempt at solving the task as a draw from a Bernoulli random variable $X$ with unknown success probability $p$. For each attempt $X_i$ we have that $X_i = 1$ if the task is solved in the $i$-th trial, and 0 otherwise. We can obtain an unbiased estimate of the solve rate $P(T^S)$ with:
\begin{align*}
    P(T^S) &\approx \widehat{P}_{t,N}\\
    \widehat{P}_{t,N} &= \frac{1}{N} \sum_{i=1}^{N} X_i 
\end{align*}
Where $\widehat{P}_{t,N}$ is a random variable corresponding to the estimate of the task solve rate with $N$ samples. Since the variance of a Bernoulli random variable is $p(1-p)$, the total variance of this estimate is:
\begin{align*}
    Var \left( \widehat{P}_{t,N} \right) &= \frac{1}{N^2}Var(\sum_{i=1}^{N} X_i)\\
    &=\frac{1}{N^2}\left(N p(1-p)\right)&\text{by independence of trials}\\
    &=\frac{p(1-p)}{N}
\end{align*}
We now consider what happens if we decompose our initial task into an arbitrary number of milestones, who each are modelled as its own Bernoulli variable $M^j$ with succes rate $p_j$. Because these milestones are subparts of the original task, we have that the original solve rate is equal to the product of the new solve rates $p=\prod_j^n p_j$. By multiplying our estimate of the solve of each new milestone, we get an overall estimate of our solve rate:
\begin{align*}
    P(T^S) &\approx \prod_j^n \widehat{P}_{m_i,N}\\
    \widehat{P}_{m_i,N} &= \frac{1}{N} \sum_{i=1}^{N} M^j_i
\end{align*}
Where $\widehat{P}_{m_i,N}$ is a random variable corresponding to the estimate of the $i$-th milestone solve rate with $N$ samples, and $M^j_i = 1$ if the milestone $j$ is solved in the $i$-th trial, and 0 otherwise. We can now use the formula for calculating the variance of a product of independent variables:
\begin{align*}
   \operatorname{Var}(X_1\cdot X_2\cdot \ldots \cdot X_n)=\left(\mathrm{E}\left(\mathrm{X}_1 \right) E\left(X_2\right) \ldots E\left(\mathrm{X}_{\mathrm{n}}\right)\right)^2 \cdot  \sum_i^n \frac{\operatorname{Var}\left(\mathrm{X}_{\mathrm{i}}\right)}{\left[\mathrm{E}\left(\mathrm{X}_{\mathrm{i}}\right)\right]^2}
\end{align*}
On our milestone estimate this gives the variance:
\begin{align*}
    \operatorname{Var}\left(\prod_j^n \widehat{P}_{m_i,N}\right)=(p_1 \cdot p_2 \cdot \ldots  \cdot p_n)^2 \cdot \sum_i^n \frac{\left(\frac{p_i(1-p_i)}{N}\right)}{p_i^2}
\end{align*}
We will now show that the variance of the milestone estimate is always smaller than or equal the end-to-end estimate:
\begin{align*}
\text{Var}\left(\prod_{i=1}^{n}{\widehat{P}_{m_i, S}}\right) &\le \text{Var}\left(\widehat{P}_{t,S}\right) \\
    (p_1 \cdot p_2 \cdot \ldots  \cdot p_n)^2 \cdot \sum_i^n \frac{\left(\frac{p_i(1-p_i)}{N}\right)}{p_i^2} &\leq \frac{p(1-p)}{N}\\
    p^2 \sum_i^n \frac{\frac{1}{p_i} - 1}{N} &\leq \frac{p(1-p)}{N}\\
    \sum_i^n \left( \frac{1}{p_i} - 1\right) &\leq \frac{1}{p}-1\\
    \sum_i^n  \frac{1}{p_i} -n + 1 &\leq \prod_i^n \frac{1}{p_i}
\end{align*}
Where the final inequality holds true, as it is a generalization of \href{https://en.wikipedia.org/wiki/Bernoulli%27s_inequality#Generalization_of_base}{Bernoulli's inequality}.

\section{Expert Bit Calibration}
\label{app:expert_bits}
The reason why we don't expect that there exists any general way of mapping expert bits into a well-calibrated probability of solving the task is that regardless of how easy a step is, the agent will be penalized at least 1 bit of expert help. This is because choosing the first index gives a penalty of $-\log_2 (\frac{1}{1+1}) = 1$ bits when using the prior from \citet{phuong2024}. Based on this, one should be able to create two tasks —an easy one and a hard one— with the same bit count, which precludes the existence of a mapping function from bits to the end-to-end solve rate.

To see why, consider the following. Imagine we have an easy task with a high solve rate and low bit count and a hard task with a low solve rate and high bit count. Now, imagine slightly altering the easy task by repeatedly adding trivial steps at the end. For example, after finding a bug in a codebase, the model must also print ``hello world'' in the console. This additional step increases the total number of expert help bits by 1 without significantly affecting the overall end-to-end probability. Since all the additional steps are trivial, the overall solve rate of the modified task remains largely unchanged. However, at some point, its bit count matches that of the hard task.

This creates a problem for one's mapping. One now has two tasks with the same bit count but vastly different solve rates. Since one cannot map from the same bit count to two different solve rates simultaneously, we conclude that no such mapping exists. While a different prior might slightly alleviate this issue, we do not see it as entirely fixable.

\section{Expert Best-of-N Cost}
\label{app:expert_cost}
The reasoning behind the index $i$ of the chosen action (counting from one) being equated to a cost of $\log_2(i(i+1))$ bits, stems from the concept of Shannon entropy. Shannon Entropy states that given a prior probability distribution over a set, where each element has a prior probability $p_i$, one can, on average, encode an item from that set using $-\log_2(p_i)$ bits. To convert these bits to success probabilities, you simply use the transformation $P(T^S) = 2^{-\text{bits}}$.

In this case, they use the prior $\frac{1}{i(i+1)}$ over the set of $N$ continuations made by the model. This prior can be a reasonable choice because it sums to 1 in the limit of large $N$:
\begin{align*}
    \sum_{i=1}^{\infty} \frac{1}{i(i+1)} &= 1
\end{align*}
Additionally, it is a decreasing function of $i$, which means that the expert is providing more information when they choose a less likely continuation. It also spreads out the probability mass more effectively than alternative priors, such as $\frac{1}{2^i}$.

\end{document}